\documentclass[10pt,twocolumn,letterpaper]{article}

\usepackage{wacv}              

\definecolor{wacvblue}{rgb}{0.21,0.49,0.74}
\usepackage[pagebackref,breaklinks,colorlinks,allcolors=wacvblue]{hyperref}

\usepackage{booktabs}       
\usepackage{amsfonts}       
\usepackage{nicefrac}       
\usepackage{microtype}      
\usepackage{makecell}
\usepackage{caption}
\usepackage{colortbl}
\usepackage{amsthm,amsmath,amssymb}
\usepackage{mathrsfs}
\usepackage{subcaption} 
\usepackage{multirow} 
\usepackage{pifont}       
\usepackage{bbding}       
\usepackage{fontawesome}  
\usepackage{xcolor}         
\usepackage{url}            
\usepackage[linesnumbered,ruled,vlined]{algorithm2e}
\SetKwInput{KwInput}{Input}                
\SetKwInput{KwOutput}{Output}              

\usepackage{graphicx}

\graphicspath{{figures/}}
\def\wacvPaperID{3248} 
\def\confName{WACV}
\def\confYear{2026}

\definecolor{commentGreen}{rgb}{0,0.5,0.05}

\definecolor{ignorecolor}{rgb}{0.875,0.875,0.75}

\newcommand{\xmark}{\textcolor[HTML]{e74c3c}{\ding{55}}}
\newcommand{\cmark}{\textcolor{commentGreen}{\bf \ding{51}}}

\title{Align Video Diffusion Model with Online Video-Centric Preference Optimization}

\author{
    Jiacheng Zhang$^1$\footnotemark[1]\qquad 
    Jie Wu$^2$\footnotemark[1]\qquad
    Weifeng Chen$^2$\qquad
    Yatai Ji$^{1}$\\
    Xuefeng Xiao$^2$\qquad
    Weilin Huang$^2$\qquad
    Kai Han$^1$\footnotemark[3]\\[0.6em]
    $^1$The University of Hong Kong\qquad $^2$ByteDance\\[0.6em]
    Project Page: \url{https://visual-ai.github.io/onlinevpo/}
}

\begin{document}
\maketitle
{
\renewcommand{\thefootnote}{\fnsymbol{footnote}}
\footnotetext[1]{Equal contribution.}
\footnotetext[3]{Corresponding Author.}
}
\begin{abstract}
Video diffusion models (VDMs) have demonstrated remarkable capabilities in text-to-video (T2V) generation. Despite their success, VDMs still suffer from degraded image quality and flickering artifacts. To address these issues, some approaches have introduced preference learning to exploit human feedback to enhance the video generation. However, these methods primarily adopt the routine in the image domain without an in-depth investigation into video-specific preference optimization. In this paper, we reexamine the design of the video preference learning from two key aspects: \textit{feedback source} and \textit{feedback tuning methodology}, and present \textbf{OnlineVPO}, a more efficient preference learning framework tailored specifically for VDMs. On the feedback source, we found that the image-level reward model commonly used in existing methods fails to provide a human-aligned video preference signal due to the modality gap. In contrast, video quality assessment (VQA) models show superior alignment with human perception of video quality. Building on this insight, we propose leveraging VQA models as a proxy of humans to provide more modality-aligned feedback for VDMs. Regarding the preference tuning methodology, we introduce an online DPO algorithm tailored for VDMs. It not only enjoys the benefits of superior scalability in optimizing videos with higher resolution and longer duration compared with the existing method, but also mitigates the insufficient optimization issue caused by off-policy learning via online preference generation and curriculum preference update designs. Extensive experiments on the open-source video-diffusion model demonstrate OnlineVPO as a simple yet effective and, more importantly, scalable preference learning algorithm for video diffusion models.
\end{abstract}
    
\section{Introduction}

Recently, Reinforcement Learning with Human Feedback (RLHF)~\cite{rlhf,bai2022constitutional} has gained significant attention for its remarkable success in enhancing the performance of large language models (LLMs)~\cite{achiam2023gpt,touvron2023llama,touvron2023llama2}. Serving as a post-training technique, it improves the generation capability of LLM via aligning the model with human preferences. 
Specifically, RLHF curates the human-annotated preference data to train a reward model (RM) to capture the human preference. Subsequently, the LLM is fine-tuned with the feedback on the LLM outputs provided by the RM via the RL algorithm, like Proximal Policy Optimization (PPO)~\cite{ppo} to guide the model to generate the human-preferred content. Building on the success of RLHF, numerous improved approaches~\cite{ipo,kto,dpo,raft,rlaif,apo} have been proposed. Among these, Direct Preference Optimization (DPO)~\cite{dpo} reformulates preference alignment as a maximum likelihood estimation problem, eliminating the annoying RL-based tuning and significantly simplifying the preference optimization pipeline. 

\begin{figure}
    \centering
    \setlength{\abovecaptionskip}{0.1cm}
    \includegraphics[width=0.5\textwidth]{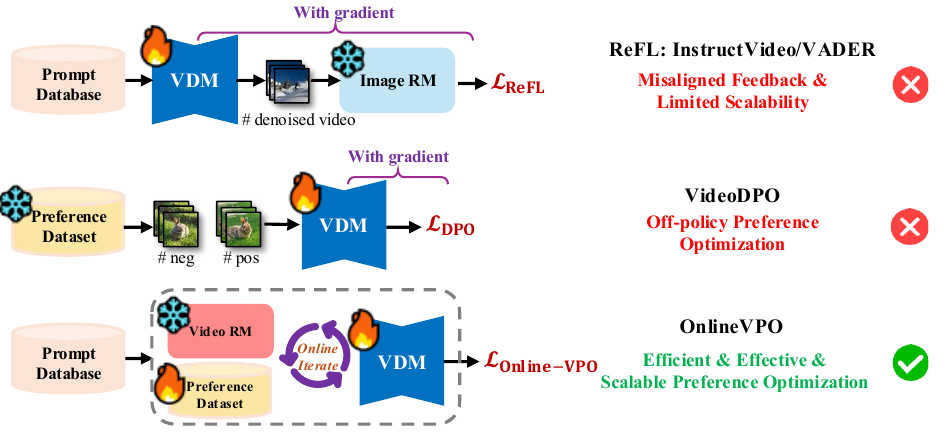}
    \caption{\textbf{Preference optimization paradigms for VDMs}. Existing methods~\cite{instructvideo,vader} suffer from misaligned feedback, limited scalability, and off-policy optimization.}
    \label{fig:teaser}
    \vspace{-0.3cm}
\end{figure}

The success of RLHF in the LLM domain has sparked growing interest in exploring preference optimization for video generation, leading to some preliminary works such as InstructVideo~\cite{instructvideo} and VADER~\cite{vader} to enhance the video generation performance via human feedback. However, upon closer examination of these approaches—particularly in terms of \textit{feedback sources} and \textit{preference fine-tuning methodologies}, we found there are still several critical limitations persisting in existing methods, as illustrated in Fig.\ref{fig:teaser}.

\begin{itemize}
\item \textit{\textbf{Modality Misaligned Preference Feedback Signal}}: Given the labor-intensive nature of video preference data collection, existing methods predominantly exploit image-level reward models~\cite{imagereward,hps,hpsv2} for video preference optimization. For instance, InstructVideo employs HPSv2~\cite{hpsv2} to obtain the frame-wise reward and derive the video feedback by weighted average. Similarly, VADER utilizes the feedback from the ImageReward~\cite{imagereward} to guide the video optimization. Since there is a domain gap between image generation and video generation, such image-level reward models may provide biased assessments due to their inability to perceive the temporal dynamics, which is a crucial aspect of video quality.

\item \textit{\textbf{Inefficient and Limited Scalable Feedback Learning}}: The existing approaches primarily rely on the Reward Feedback Learning (ReFL) framework~\cite{imagereward} to fine-tune the VDMs, which involves scoring the entire video output decoded by VAE using the reward model and propagating the feedback signal through the reward model to the sampling chain, which leads to significant memory consumption as depicted in Fig.\ref{fig:gpu}. Consequently, it is challenging to scale these methods to support larger-parameter VDMs/reward models or higher-resolution/longer-duration video. Additionally,  the ReFL framework is evidenced to be prone to suffer reward hacking issues~\cite{rewardhack}. 
\end{itemize}

In response to these issues, we introduce \textit{\textbf{OnlineVPO}}, a low-cost, effective, and scalable video preference optimization framework. Specifically, we first delve into the several video feedback strategies commonly used in the existing methods. Our results reveal that the image-level reward model exhibits unsatisfactory alignment performance. To seek a low-cost alternative to human feedback, we propose to exploit the video assessment model (VQA)~\cite{videoscore,qalign,t2vqa} with rich domain-specific knowledge to provide the modality-aligned video feedback. As for the feedback learning methodology, instead of adopting the ReFL framework as the previous practice, we explore the more scalable DPO paradigm for video preference optimization. However, we found that it is not trivial to apply DPO to video generation as it suffers from insufficient optimization. Specifically, DPO typically fine-tunes with a preference dataset collected ahead of training. This preference feedback is typically purely offline due to the current policy's inability to receive precise feedback about its generations during training. This engenders a notable distribution disparity between the policy from which preference pairs are collected and the policy to align. To tackle the issue, we develop an online DPO algorithm for video preference optimization. Specifically, instead of aligning VDMs toward the pre-collected preference pair, we exploit the VQA model to derive the feedback on the latest generated video sample from the current video generation policy model on the fly, which achieves more effective on-policy preference optimization. Additionally, we design a curriculum reference update strategy to dynamically update the reference model to further enhance the optimization efficiency. Extensive experiments on the open-sourced VDMs demonstrate that OnlineVPO achieves superior effectiveness and scalability in preference optimization. Furthermore, for the first time, we provide a systematic investigation into key aspects of video preference optimization, including reward learning efficiency, memory scalability, \etc. We hope that our work will inspire and facilitate further exploration in this field.

The main contributions of our work are as follows:
\begin{itemize}
\item \underline{\textit{More Efficient Video Feedback Sources}}. Through careful quantitative study, we reveal that the common practice utilizing image-level RM fails to provide human-aligned video preference feedback, while the VQA models can serve as a superior alternative.

\item \underline{\textit{Improved Preference Learning Methodology}}. We develop OnlineVPO, a more effective and scalable video preference optimization algorithm for VDMs with the online preference sample generation and curriculum reference update design.

\item \underline{\textit{New Insights for Video Preference Learning}}. We conduct extensive experiments to demonstrate the superiority of our method and provide the first systematic investigation of preference optimization of VDMs, which facilitates future progress in this area.
\end{itemize}
\label{sec:formatting}
\subsection{Diffusion-based Video Generation}
Diffusion models~\cite{ddpm,ldm,chen2023pixart} have shown impressive performance in generating high-quality videos. Early works~\cite{animatediff,magicvideo,lavie,modelscope,alignvideo,tuneavideo,makeavideo,text2video} commonly inflate the pre-trained image diffusion model~\cite{ldm} into the video diffusion model by incorporating temporal layers(\eg, temporal convolution or temporal attention). Subsequently, large video pre-trained diffusion models~\cite{videocrafter1,videocrafter2,svd,lumiere} represented works such as VideoCrafter~\cite{videocrafter1} exhibit impressive video quality with large-scale video pre-training datasets. However, these methods encounter limitations in generating long videos due to inherent limitations in capacity and scalability within the UNet design. Spearheaded by groundbreaking works like Sora~\cite{sora},  a wave of DiT-based video diffusion models~\cite{easyanimate, opensora,vidu,cogvideo,cogvideox} has steadily emerged. With the large-scale training and scalability of DiT~\cite{dit,chen2023pixart} architecture, these models can generate longer videos of up to 1 minute. Despite these advancements, challenges such as poor image quality, frame flicker, and subject inconsistencies in the generated video persist. In this study, we aim to address these issues through preference optimization.
\subsection{Preference Learning from Human Feedback}
Recently, the preference optimization~\cite{ipo,kto,dpo,raft,rlaif,apo} represented by Reinforcement Learning with Human Feedback (RLHF) has emerged as a standard technique in modern LLMs~\cite{achiam2023gpt,llama3}. Inspired by their success, there is a growing interest in exploring preference optimization in other domains. For image generation, various preference optimization approaches have been developed~\cite{hps,diffusion_dpo,hpsv2,ddpo,d3po,dpok,aligndiff,imagereward}. For example, DiffusionDPO~\cite{diffusion_dpo} extends the idea of direct preference optimization~\cite{dpo} to image generation by adapting it to the diffusion training objective. Recently, Reward Feedback Learning (ReFL)~\cite{imagereward} has emerged as an effective image preference optimization method. Specifically, ReFL utilizes a tailor-designed image-based reward model, namely, ImageReward, to fine-tune the image diffusion model via sampling chain gradient propagation. Inspired by these, there has also been some work~\cite{instructvideo,t2v-turbo,vader} to explore the preference optimization for video generation. For example, InstructVideo~\cite{instructvideo} proposes to utilize the image reward model HPSv2~\cite{hpsv2} to score the sampled frames and then employ a frame-wise weighting strategy to obtain video rewards for preference fine-tuning. In addition to employing ImageReward to provide feedback, VADER~\cite{vader} also explores video self-supervised models like V-JEPA~\cite{v-jepa} as a reward function to facilitate the temporal coherence. However, neither the image-level reward model nor the self-supervised model, neither of which is specifically trained to provide feedback about video quality, especially for the temporal dynamics. In our work, we reveal that the commonly used image-level reward model fails to distinguish human-preferred videos. Besides, most of these preliminary efforts adopt the ReFL framework to achieve feedback tuning, which is prone to suffer from the memory issue and reward hack, while leaving the DPO-based video preference optimization under-explored. In this paper, we make a pioneering systematic exploration of DPO paradigms in the video generation domain. Note that concurrent to our work, VideoDPO~\cite{videodpo} also explores the adaptation of DPO for video generation. However, they fail to solve the inefficient learning issue of DPO caused by the offline optimization.
\begin{table*}[h]
    \centering
    \small
    
    \setlength{\tabcolsep}{-1pt} %
    \setlength{\arraycolsep}{-1pt} %
    \setlength{\abovecaptionskip}{0.1cm}
   \begin{tabular}{@{}p{4cm}<{\centering} | @{}p{2.1cm}<{\centering}@{} p{2.1cm}<{\centering}@{} p{3.8cm}<{\centering}@{} | @{}p{2.4cm}<{\centering}@{} p{2.2cm}<{\centering}@{}}
    \hline
   \multirow{2}{*}{Method} 
    &
      \multicolumn{3}{c}{\textbf{Reward Model}} &
      \multicolumn{2}{c}{\textbf{Feedback Attributes}}  \\ \cline{2-6} 
   & RM Guidance  & Video Reward &  Video Quality-Aware &  Online Feedback  & Scalability \\
    \hline
    InstructVideo & \cmark  & \textcolor{red}{\xmark} & \textcolor{red}{\xmark} & \textcolor{green}{\cmark} & \textcolor{red}{\xmark}  \\
    T2V-Turbo & \textcolor{green}{\cmark}  & \textcolor{green}{\cmark} & \textcolor{red}{\xmark}  & \textcolor{green}{\cmark} & \textcolor{red}{\xmark}   \\
    VADER & \textcolor{green}{\cmark}  & \textcolor{green}{\cmark} & \textcolor{red}{\xmark}  & \textcolor{green}{\cmark} & \textcolor{red}{\xmark}  \\
    VideoDPO  & \textcolor{green}{\cmark}  & \textcolor{green}{\cmark} &  \textcolor{green}{\cmark}  & \textcolor{red}{\xmark} & \textcolor{green}{\cmark}  \\
    \rowcolor{gray!10} \textbf{OnlineVPO(Ours)} & \textcolor{green}{\cmark}   & \textcolor{green}{\cmark}  & \textcolor{green}{\cmark}   & \textcolor{green}{\cmark}  & \textcolor{green}{\cmark}    \\
    \hline
\end{tabular}
  \caption{\textbf{Comprehensive Comparison of Video Preference Optimization Methods}. We compare different preference learning methods of VDMs regarding the reward model and feedback attribute.}
    \label{tab:comparison_method}
    \vspace{-0.5cm}
\end{table*}

\section{Methodology}
\begin{figure*}[t]
    \centering
    \setlength{\abovecaptionskip}{0.1cm}
    \begin{minipage}[t]{0.6\textwidth}
        \centering
        \includegraphics[width=\linewidth]{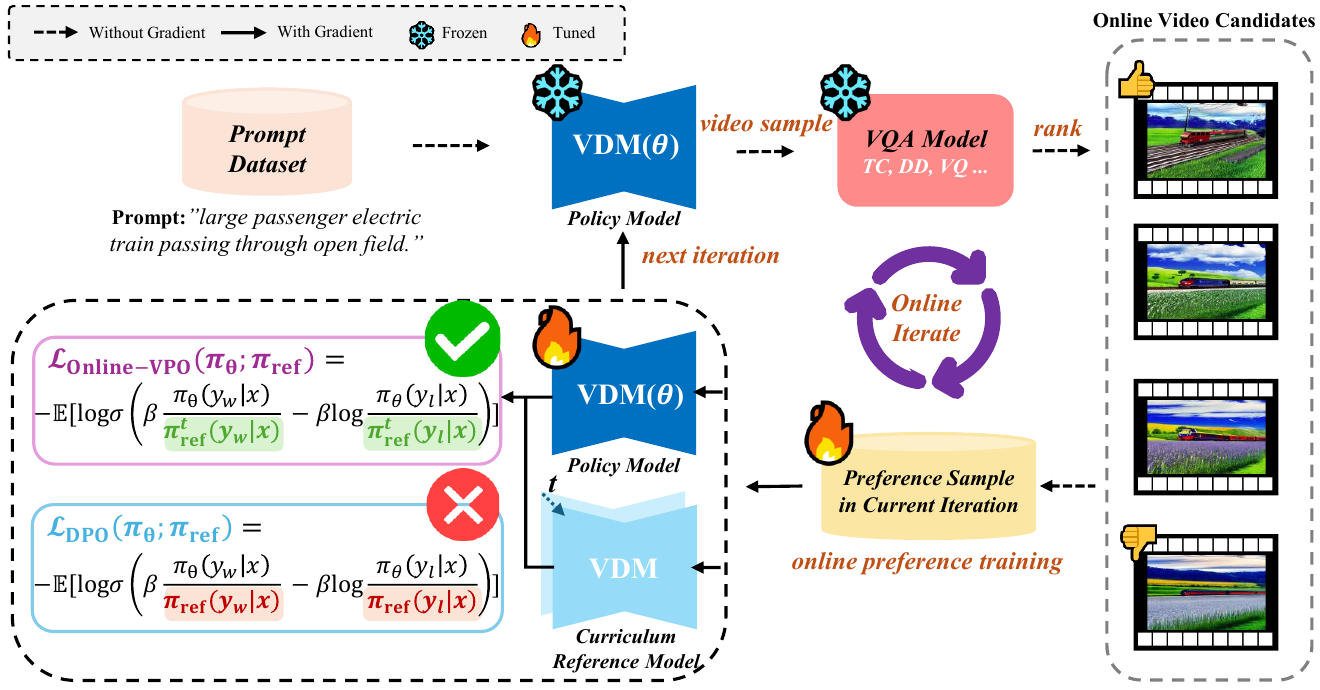}
        \caption{\textbf{OnlineVPO Pipeline}. OnlineVPO imposes the online video preference optimization with the \textit{preference samples generated on-the-fly} with the guidance of video-centric reward and achieves efficient learning by updating the reference model in a curriculum manner.}
        \label{fig:pipeline}
    \end{minipage}
    \hfill
    \begin{minipage}[t]{0.38\textwidth}
        \centering
        \includegraphics[width=0.95\linewidth,height=0.9\textwidth]{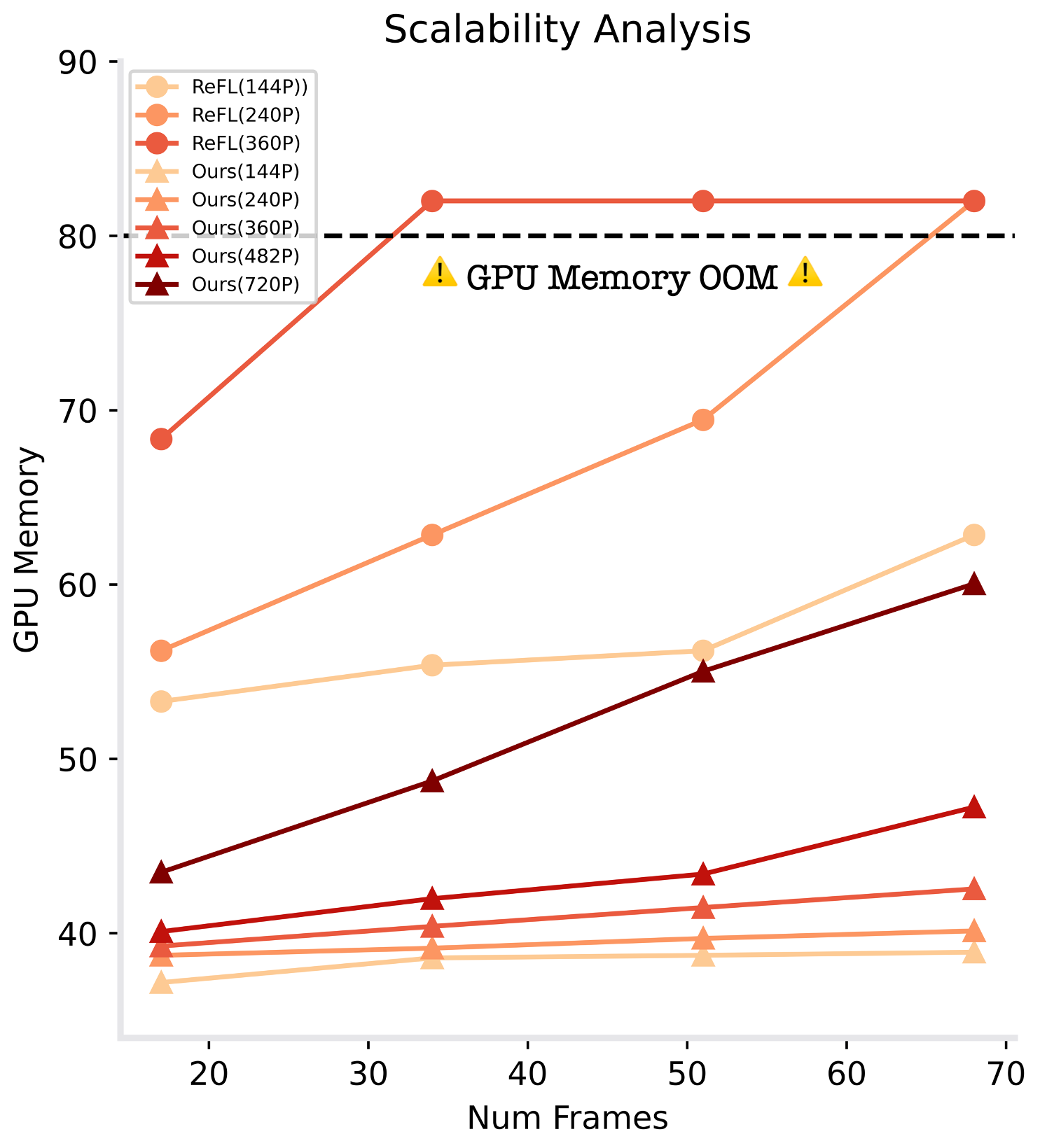}
        \caption{The \textbf{Scalability Analysis} between OnlineVPO and ReFL. The dotted line is the limit of A100 GPU Memory.}
        \label{fig:gpu}
    \end{minipage}
    \vspace{-0.5cm}
\end{figure*}

In this section, we introduce OnlineVPO for preference optimization of VDMs. Firstly, we present some preliminaries regarding two common RL-free preference learning paradigms, namely, ReFL and DPO. Subsequently, we present our investigation into the video feedback and the online preference optimization method. The overview of our method is depicted in Fig.\ref{fig:pipeline}.

\subsection{Preliminary}
\noindent\textbf{Reward Feedback Learning (ReFL).} ReFL~\cite{imagereward} is originally proposed for image diffusion models. Specifically, it begins with an input prompt $c$, initializing a latent variable $x_T$ at random. The latent variable $x_T$ is then progressively denoised without gradient until reaching a randomly selected timestep $t$, \ie $x_{T} \rightarrow x_{T-1} \cdots x_t$. Subsequently, with gradient turns on, the denoised image $x^{\prime}_0$ is directly predicted from $x_t$ and decoded, \ie $x_t \rightarrow x^{\prime}_0$. A human-preference aligned reward model $r$ is then applied to score this denoised image, generating the expected preference score $r_{\theta}( x^{\prime}_0, c)$. ReFL aligns the diffusion model with human preference by directly maximizing the preference scores of the generated image as follows:
\begin{equation}\label{eq:refl}
\mathcal L_{\mathrm{ReFL}}(\theta) = \mathbb E_{c \sim p(c)}\mathbb E_{x^{\prime}_0 \sim p(x^{\prime}_0|c)}[-r(x^{\prime}_0,c)]
\end{equation}
It is straightforward to apply ReFL to VDMs as in \cite{instructvideo,vader}.

\noindent\textbf{Direct Preference Optimization (DPO).} DPO~\cite{dpo} is one of the most popular preference optimization alternatives to RL-based RLHF. Instead of learning an explicit reward model, DPO reparameterizes the reward function $r$ using a closed-form expression with the optimal policy:
\begin{equation}\label{eq:dpo_policy}
r(x, y)=\beta \log \frac{\pi_{\theta}(y \mid x)}{\pi_{\mathrm{ref}}(y \mid x)}+\beta \log Z(x)
\end{equation}
where $\pi_{\theta}$ and $\pi_{\text{ref}}$ is the policy model and reference model. $x$ and $y$ are the prompt and response.
DPO formulates the probability of preference generation via the policy model rather than the reward model, yielding the objective as follows:
\begin{footnotesize} 
\begin{align}
\label{dpo_objective}
&\mathcal{L}_{\mathrm{DPO}}(\theta) = \\
&\mathbb{E}_{(y_w, y_l) \sim D}\left[\log \sigma\left(\beta \log \frac{\pi_{\theta}\left(y_{w} \mid x\right)}{\pi_{\mathrm{ref}\left(y_{w} \mid x\right)}}-\beta\log \frac{\pi_{\theta}\left(y_{l} \mid x\right)}{\pi_{\mathrm{ref}}\left(y_{l} \mid x\right)}\right)\right]
\end{align}
\end{footnotesize}
where $(y_w, y_l)$ is the preferred and unpreferred response from the preference dataset $D$.

\begin{table}[]
    \centering
    \footnotesize
    \setlength{\abovecaptionskip}{0.1cm}
    \tabcolsep=3.5pt
    \begin{tabular}{c|c|ccc}
    \hline { \makecell{Feedback \\ Source }}& \begin{tabular}{c} 
    Preference \\
    MRR (\%)
    \end{tabular}$\uparrow$ &\begin{tabular}{c} \\$@1$\end{tabular} & \begin{tabular}{c} 
    Recall (\%)$\uparrow$ \\
    $@2$
    \end{tabular} & \begin{tabular}{c} \\$@4$\end{tabular} \\
    \hline Aesthetic Score & 42.73 & 12.45 & 26.24 & 77.37  \\
    ImageReward & 44.17 & 14.25 & 27.44 & 78.69  \\
    MPS & 45.52 & 15.06 & 27.39 & 79.45 \\
    HPSv2 & 46.30 & 24.60 & 31.09 & 80.12  \\
    \hline VideoScore & $\mathbf{67.23}$ & $\mathbf{41.38}$ & $\mathbf{62.74}$ & $\mathbf{92.66}$ \\ 
    \hline
    \end{tabular}
    \caption{\textbf{Video feedback source analysis} of several commonly used image-level reward models on distinguishing the human-preferred video from the candidates. (The temporal dimension of VideoScore is used.)}
    \label{rm_analysis}
    \vspace{-0.6cm}
\end{table}

\subsection{Investigation on Video Preference Feedback}\label{sec:analysis}
Feedback source is the most crucial part of preference optimization, determining the optimization outcome's upper bound. Existing methods~\cite{instructvideo,t2v-turbo} employ various image-level reward models to provide the feedback signal for video generation. We want to ask \textit{whether these feedback strategies work appropriately like humans?} To answer this question,  we curate a comprehensive prompt set that covers various scenes, objects, and events, and generate 8 candidate videos with several state-of-the-art video diffusion models. Then, 5 proficient human annotators are asked to select the best video via voting. After that, we treat these annotations as our ground truth preference and utilize several commonly used image models to rank these candidates. To quantify the preference feedback performance, we define the preference mean reciprocal rank (MRR):
\begin{equation}\label{eq:dpo_policy}
\text{MRR}=\frac{1}{|Q|}\sum_{i=1}^{|Q|}\frac{1}{rank_i}
\end{equation}
where $|Q|$ is the number of evaluation cases and $rank_i$ denotes the ranking of the ground truth preference video within the $i$-th case's reward rank results.  We also report MRR results of Recall$@k$,  which refers to the proportion of preferred videos ranked at the top-k elements.  Apparently, higher MRR represents better feedback performance, which demonstrates better feedback alignment with human preference. The results are summarized in Tab.\ref{rm_analysis}. It is evident that the image-level preference model performs poorly at distinguishing the most preferred samples in the context of the video. These results validate the modality gap between image and video generation, revealing the challenge of obtaining reliable video feedback. Note that the most effective way to obtain the video feedback is always to train a reward model on the curated, manually annotated video preference dataset. However, collecting such a large-scale dataset is labor-intensive and expensive. Here, we focus on seeking a low-cost video feedback solution. To this end, we propose to exploit the off-the-shelf model (\eg, VideoScore~\cite{videoscore}) developed for the video quality assessment (VQA)~\cite{videoscore,qalign} task as they exhibit decent feedback alignment with humans, although they are not specifically designed for preference learning. We speculate this because it has encoded rich video domain knowledge. We then take the VQA model as an alternative video feedback source to human preference.

\subsection{Online Video Preference Optimization}
We present the first exploration of DPO in the video generation domain, and introduce OnlineVPO, an efficient and scalable online preference learning framework tailored for VDMs, which consists of two novel designs: online preference sample generation and curriculum reference update.

\noindent \textbf{Online Preference Sample Generation:} In our initial experiments (see  Tab.\ref{analysis_ablation}(d)), we observed that directly applying standard DPO to video generation failed to yield significant performance improvements and even led to notable degradation in metrics such as dynamic degree. We attribute this to the off-policy nature of DPO. Specifically, DPO aligns the model's predictions with the static preference dataset collected ahead of training, which remains fixed throughout the optimization process. As the model is continuously updated, a distributional disparity arises between the model's latest generated samples and the static preference samples used for alignment. This mismatch prevents the model from receiving precise feedback on its predictions, ultimately leading to sub-optimal performance. To address this issue, we propose online preference sample generation, which enables more effective on-policy preference optimization. Concretely, instead of relying on fixed preference pairs $(v_w, v_l)$
from a pre-collected dataset, we sample multiple videos $V=\{v_1, v_2, \cdots, v_N\}$ using the policy model $\pi_{\theta}$ at each optimization step. The preferred and unpreferred samples $(y_w, y_l)$ are then determined using the VQA model in a online manner, where $(y_w, y_l) = ( v_{\arg\max_i s_{i}}, v_{\arg\min_i{s_i}})$, where $S=\{s_1, s_2, \dots,s_N \}$ is the reward score obtained from the VQA model upon the online video candidate $v_i$.
 This mechanism allows the VDMs to receive accurate, real-time feedback on their generated samples during the alignment process, providing more effective and dynamic supervision to steer the model toward preferred outputs.

\noindent \textbf{Curriculum Preference Update:} 
Similar to standard RL-based RLHF, DPO also employs a reference model to constrain deviations in the model's output distribution. While this design stabilizes training, it inherently limits the model's exploration capacity. Through our experiments, we identified a critical phenomenon: the optimization efficiency slows significantly after a certain number of optimization steps (see Fig.\ref{fig:reward_change}). This occurs because the policy model exhausts its exploration potential within the constraints imposed by the fixed reference model, achieving a level of preference alignment that surpasses the reference model. As a result, the static reference model becomes a bottleneck, restricting the policy model from further improvement. To address this limitation, we introduce a curriculum reference update mechanism, where the reference model is periodically updated with the newly aligned policy model after a fixed number of optimization steps. This approach balances optimization efficiency with training stability, enabling the policy model to continuously explore better alignments in a curriculum manner while avoiding excessive deviation from the reference distribution.

\noindent \textbf{Discussion on Online Preference Learning.}\label{understanding}
We provide further discussion on the design of our online preference learning approach. Notably, the RL step in widely used RL-based RLHF for LLMs also operates in an online manner, as training data is acquired interactively. Specifically, PPO-based RLHF methods interact online with the language model being aligned, estimate unbiased gradients using policy gradient techniques on the generated samples, and employ a value function to reduce gradient estimation variance. This online interaction has been shown to be crucial to the success of RLHF in aligning language models, as demonstrated by \cite{ziegler2019fine}. While DPO offers a compelling alternative to PPO-based RLHF, it relies on pre-collected offline datasets, which limit its effectiveness. Our OnlineVPO bridges the gap between DPO and classical RLHF techniques by integrating online preference sample generation into the DPO optimization framework. On one hand, OnlineVPO enhances DPO's performance by leveraging dynamically generated online preference samples. On the other hand, it avoids the optimization challenges associated with complex RL tuning by utilizing the simpler DPO objective. In essence, OnlineVPO combines the strengths of both classical PPO-based RLHF methods and DPO, resulting in superior performance for aligning VDMs.
\section{Experiment}
\subsection{Experiment Setup}
\noindent \textbf{Implementation details.}\label{implement_detail} We implement our method with the two advanced open-sourced VDMs: UNet-based VideoCrafter-v2~\cite{videocrafter2} and DiT-based OpenSora v1.2~\cite{opensora}. We fine-tune these two models with WebVid-10M~\cite {webvid}. We employ AdamW~\cite{adamw} with learning rates of 1e-5 for VideoCrafter and 2e-6 for OpenSora. The batch size is set to 8. During online sample generation, we generate 6 candidate videos at 240p resolution with 30 sample steps, each comprising 34 frames with a 9:16 aspect ratio for OpenSora. For VideoCrafter, we sample the videos with 16 frames at 512$\times$320 resolution. The curriculum update interval is set to $K=200$. Unless specified otherwise, we utilize the video quality assessment model, VideoScore~\cite{videoscore} to provide video feedback. Please refer to our Appendix for more details.

\begin{table*}[h]
    \centering
    \tabcolsep=2pt
    \setlength{\abovecaptionskip}{0.1cm}
    \begin{tabular}{c  cc c c c c c c}
    \toprule Models & \textbf{Quality Score} & \begin{tabular}{cl}
    Subject \\
    Consist.
    \end{tabular} & \begin{tabular}{cl}
    Background \\
    Consist.
    \end{tabular} & \begin{tabular}{cl}Temporal \\ Flicker.\end{tabular} & \begin{tabular}{cl}
    Motion \\
    Smooth.
    \end{tabular} &  \begin{tabular}{cl}Aesthetic \\ Quality \end{tabular} & \begin{tabular}{cl}
    Dynamic \\
    Degree
    \end{tabular} & \begin{tabular}{cl}Image \\Quality\end{tabular} \\
    \midrule
      ModelScope-T2V~\cite{modelscope} & 73.70  & 86.41 &92.25 & 93.69 & 95.48 & 42.41 & 66.00 & 61.69 \\
    LaVie~\cite{lavie}  & 78.23 & 92.50 & 95.92 & 94.72 & 96.47 & 56.17 & 60.00 & 62.09 \\
     VideoCrafter1~\cite{videocrafter1}&  80.47 & 96.61 & 96.47 & 95.64 & 97.64 & 59.52 & 48.00  & 65.03 \\
     EasyAnimate~\cite{easyanimate}&  79.87 & 92.32 & 95.40 & 96.69 & 97.81& 54.82 & 81.00  & 54.63 \\
     \midrule
    InstructVideo~\cite{instructvideo}&  79.95 & 96.45 & 97.08 & 95.30 & 96.76 & 50.01 & 61.22 & 70.09 \\
    VADER~\cite{vader} & 80.02 &
    95.53 & 97.11  &97.42 & 98.89 & 53.43 & 41.12 & 66.08 \\
    VideoDPO~\cite{vader} &  81.87 & 96.98 & 97.83 & 97.45 & 98.98 & 59.23 & 45.00 & 67.33  \\
    \midrule
     OpenSora &  79.43 & 95.35 & 96.42 & 98.34 & 98.71 & 52.74 & 44.00 & 62.41 \\
    \rowcolor{gray!20}   \textit{\textbf{OpenSora + OnlineVPO}}&   \textbf{81.98}& 97.58 & 97.74 & 98.73 & 99.36 & 55.37 & 43.00 &  67.36 \\
   VideoCrafter2~\cite{videocrafter2} &80.78 & 96.65 & 97.52 & 95.75 & 97.67 & 59.10 & 50.00 & 66.85 \\
    \rowcolor{gray!20}  \textit{\textbf{ VideoCrafter2 + OnlineVPO}} &\textbf{82.91} & 97.98 & 98.24 & 97.53 & 98.85 & 62.31 & 47.00 & 68.92 \\
    
\bottomrule
\end{tabular}
 \caption{\textbf{Quantitive Performance Comparison} on VBench~\cite{huang2024vbench}. OnlineVPO demonstrates superior performance in optimizing the key aspects of video generation, such as subject consistency, temporal flicker mitigation, \etc.}
 \label{main_results}
 \vspace{-0.6cm}
\end{table*}

\noindent \textbf{Baseline Methods.} 
We compared our method with the current state-of-the-art video preference learning approaches, InstructVideo~\cite{instructvideo}, VADER~\cite{vader}, and VideoDPO~\cite{videodpo}. Both InstructVideo and VADER employ the ReFL framework to achieve preference optimization, with HPSv2~\cite{hpsv2} and PickScore~\cite{pickscore} to serve as their video feedback, respectively. VideoDPO~\cite{videodpo} first trains a video OmniScore on the curated video feedback dataset, and then fine-tunes the VDMs with the re-weighted DPO objective on the preference dataset pre-constructed via the OmniScore.

\noindent \textbf{Evaluation Metrics.} 
We employ VBench~\cite{huang2024vbench} to evaluate our method. VBench is a standard video evaluation benchmark designed to evaluate T2V models from various dimensions comprehensively. Each dimension within VBench is customized with specific prompts and evaluation methods. We choose seven representative dimensions of the video quality that align with human perception for our evaluation: dynamic degrees, subject consistency, background consistency, aesthetic quality, image quality, and motion smoothness. We primarily focus on OpenSora to demonstrate our method in the main paper, more results can be found in the Appendix.

\subsection{Compared with State-of-the-art Methods}\label{performance_comparison}
\noindent\textbf{Quantitative Comparison.} Tab.~\ref{main_results} presents a quantitative comparison between our method and existing approaches. Our method demonstrates consistent performance improvements over both OpenSora and VideoCrafter-v2. For example, integrating OnlineVPO with VideoCrafter v2 outperforms the state-of-the-art method VideoDPO. When compared to VADER, which leverages a mixed reward combining image-level rewards and video self-supervised reward functions, our method exhibits superior performance, particularly in the dimensions of subject consistency and temporal flicker. This underscores the importance of feedback that is explicitly aware of video quality. We observe that our method may result in a drop in dynamic degree. This can be attributed to an inherent trade-off between temporal consistency and dynamic degree: higher motion often compromises temporal coherence, which is also observed in \cite{t2v-turbo}. It should also be noted that blindly increasing the dynamic degree does not imply higher video quality. For instance, while EasyAnimate and ModelScope-T2V achieve higher dynamic degrees, the generated videos exhibit poor subject consistency, which affects the video-watching experience significantly and leads to an inferior overall quality score.

\noindent\textbf{User Study.} We collected 100 commonly used prompts from ~\cite{opensora,videocrafter1,videocrafter2} and then conducted a two-alternative forced choice (2AFC) test with ten volunteers given pairs of videos that were randomly sampled from the candidates of the corresponding VDMs. Each volunteer was asked to select the preferred video. Fig.\ref{fig:user} summarizes the preference rates of OnlineVPO compared to other methods. The results demonstrate that our approach consistently outperforms competing methods under human subjective evaluation.

\begin{figure*}[t]
\centering
\setlength{\abovecaptionskip}{0.1cm}
\includegraphics[width=0.99\linewidth]{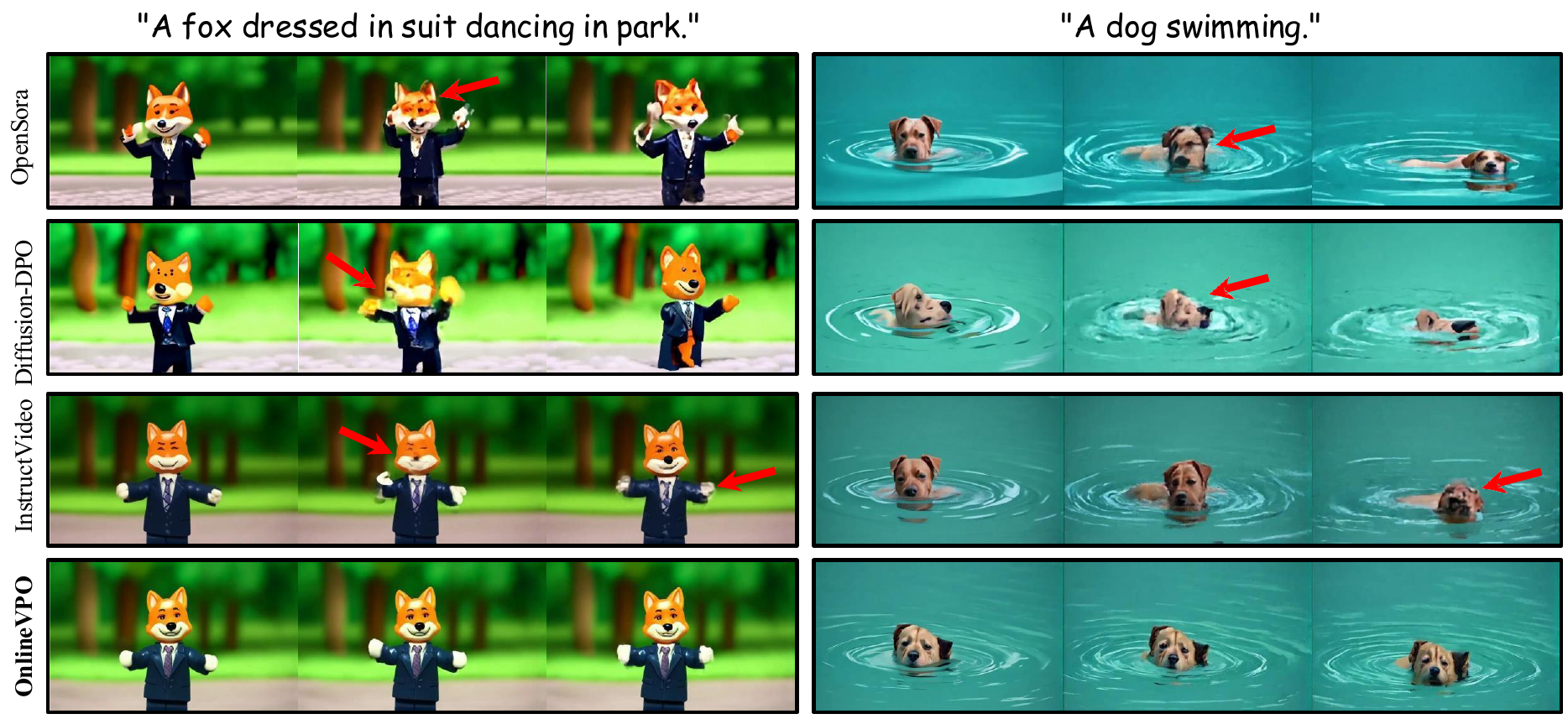}
\caption{\textbf{Visual Comparison} between different preference optimization methods upon OpenSora. The proposed OnlineVPO exhibits better frame quality and higher temporal consistency than other methods.}
\label{fig:vis}
\vspace{-0.6cm}
\end{figure*}

\noindent \textbf{Visualization.} 
We showcase some visual comparison results between ours and other methods based on OpenSora in Fig.\ref{fig:vis}. Since VideoDPO only releases the code for VideoCrafter v2, to facilitate the understanding of the superiority of our method over the naive DPO scheme, we directly applied the DPO to OpenSora and termed it as \textit{Diffusion-DPO}. It can be observed that the videos generated by our model are more stable and less susceptible to temporal collapse. For instance, in the first example,  the fox generated by both the original OpenSora and Diffusion-DPO exhibits collapse, resulting in distorted and broken facial features. InstructVideo shows some better results, but still suffers from blurred frames, such as the blurry face and distorted limbs of the fox. In contrast, our results achieve higher image quality, greater clarity, and improved temporal consistency across the entire sequence.
\subsection{Ablation Study}
We conduct a series of thorough ablation studies and endeavor to provide some insights into preference learning for the video generation domain. Unless otherwise stated, our experiments are performed with Open-Sora v1.2 as the VDMs and VideoScore as the video reward model.

\noindent\textbf{Video Reward Model Benefits More.} We investigate the effectiveness of different feedback sources in the context of video generation in Tab.\ref{analysis_ablation}(a). Specifically, we explore several popular image-level reward models for video preference learning, including Laion Aesthetic Predictor\footnote{https://github.com/christophschuhmann/improved-aesthetic-predictor}, ImageReward~\cite{imagereward}, HPSv2~\cite{hpsv2}, and MPS~\cite{mps}. Additionally, we examine the potential of video quality assessment (VQA) models, such as Q-Align~\cite{qalign} and VideoScore~\cite{videoscore}, to provide video-level feedback.  We exploit these models to construct online preference samples in OnlineVPO. We observe that leveraging image-level reward models as feedback sources yields certain performance improvements. However, these enhancements are primarily limited to image-level attributes, such as image quality and aesthetic quality, with marginal gains in video-specific dimensions like dynamic degree and temporal consistency. In contrast, VQA models, which possess a comprehensive understanding of video quality, result in consistent performance improvements across various dimensions after preference tuning, with VideoScore achieving optimal performance. These findings underscore the importance of modality-aligned, video domain-specific feedback and also validate our analysis in Sec.\ref{sec:analysis}.

\noindent \textbf{OnlineVPO or ReFL.} We evaluate the effectiveness of ReFL, another mainstream reward optimization paradigm, in the context of video generation. ReFL requires the reward model to be differentiable to enable gradient propagation from the reward back to the diffusion model. Since the VideoScore used in our work is an LLM-based differentiable VQA model, we take it as the reward model in ReFL and perform the reward-feedback tuning following the practice in \cite{imagereward,instructvideo}. It can be observed from Tab.\ref{analysis_ablation}(c) that ReFL achieves inferior optimization performance even when using the same video reward. Furthermore, unlike ReFL, OnlineVPO imposes no constraints on the differentiability of the reward model, allowing for greater flexibility in employing diverse video feedback strategies.

\noindent \textbf{Optimization Dimension.} We analyze the performance of video preference optimization across different preference dimensions. Thanks to the video quality aware VQA model, we can easily obtain the assessment of the generated video quality from different aspects, including  \textit{visual quality}, \textit{temporal consistency}, \textit{dynamics}, and \textit{t2v alignment}.  we also investigated the use of feedback from all dimensions for optimization (termed as \textit{global}),  where online preference pairs are constructed by averaging reward scores across all dimensions. The results are summarized in Tab.\ref{analysis_ablation}(b).  Interestingly, we found: (1) Feedback focusing on temporal consistency yields the best optimization performance, underscoring the importance of temporal-aware feedback in video preference optimization. Notably, optimizing for temporal consistency also enhances performance in other dimensions (\eg, aesthetic quality), whereas feedback from other dimensions does not exhibit this cross-dimensional benefit.  (2) The performance with the feedback from all dimensions is inferior to the feedback on a particular dimension (\eg, temporal consistency). We hypothesize that this is due to potential conflicts arising from integrating feedback across different dimensions. For instance, high motion intensity may introduce blurring or artifacts, which harm the aesthetic quality. while static frames may excel in aesthetic quality and favor high temporal consistency, yet their motion amplitude can be severely affected. Averaging feedback across these conflicting dimensions can result in contradictory signals, ultimately hindering optimization.

\begin{figure*}
    \centering
    \setlength{\abovecaptionskip}{0.1cm}
    \begin{subfigure}{0.49\textwidth}
      \centering
      \footnotesize
        \tabcolsep=3pt
       \begin{tabular}{c| c c c }
     
       Method & \begin{tabular}{cl} Dynamic \\ Degree \end{tabular} &\begin{tabular}{cl}  Subject \\ Consist. \end{tabular}&\begin{tabular}{cl} Aesthetic \\ Quality \end{tabular}\\
        \hline 
        Aesthetic  & 40.1 & 95.04 & 52.33 \\
         ImageReward & \textbf{46.0} & 96.26 & 51.04 \\
        MPS & 42.0 & 95.93 &  51.54\\
        QAlign & 43.6 & 96.36 & 53.75 \\
    \rowcolor{gray!10}   VideoScore &43.0 &  \textbf{97.58} &\textbf{ 55.37}  \\
    
    \end{tabular}
      \caption{OnlineVPO with different video feedback sources. }
        \label{tab:policy_rm}
        
    \end{subfigure}
    \hfill
    \begin{subfigure}{0.49\textwidth}
        \centering
        \footnotesize
        \tabcolsep=3pt
        \begin{tabular}{c| c c c }
        Feedback Dimension & \begin{tabular}{cl} Dynamic \\ Degree \end{tabular} &\begin{tabular}{cl}  Subject \\ Consist. \end{tabular}&\begin{tabular}{cl} Aesthetic \\ Quality \end{tabular} \\
            \hline 
            Visual quality & 38.0 & \underline{96.56} & \underline{52.97}  \\
            Dynamic degree & \textbf{57.0} & 92.09 & 49.02 \\
            T2V alignment & 44.0 & 96.19 & 50.52 \\
             \rowcolor{gray!10} Temporal consistency & 43.0 & \textbf{97.58} & \textbf{55.37} \\
            Global & \underline{55.0} & 95.37 & 51.33 \\
            
        \end{tabular}
        \caption{Different feedback dimensions with VideoScore.}
        \label{tab:dimension}
        
    \end{subfigure}

    \medskip

    \begin{subfigure}{0.49\textwidth}
        \centering
        \footnotesize
        \tabcolsep=3pt
       \begin{tabular}{c| c c c c }
       Method & \begin{tabular}{cl} Dynamic \\ Degree \end{tabular} &\begin{tabular}{cl}  Subject \\ Consist. \end{tabular}&\begin{tabular}{cl} Aesthetic \\ Quality \end{tabular} &\begin{tabular}{cl} Image \\ Quality \end{tabular}\\
        \hline 
        ReFL & 40.0 & 96.59 & 51.84 & 63.06 \\
         \rowcolor{gray!10} Ours & \textbf{43.0} & \textbf{97.58} &\textbf{ 55.37} &\textbf{67.36}\\

    \end{tabular}
      \caption{OnlineVPO and ReFL with VideoScore (w/ TC).}
        \label{tab:dimension}
        
    \end{subfigure}
    \hfill
    \begin{subfigure}{0.49\textwidth}
        \centering
        \footnotesize  
       \begin{tabular}{c| c c c }
       Method & \begin{tabular}{cl} Dynamic \\ Degree \end{tabular} &\begin{tabular}{cl}  Subject \\ Consist. \end{tabular}&\begin{tabular}{cl} Aesthetic \\ Quality \end{tabular}\\
        \hline 
        Offline-policy & 28.0 & 97.07 & 53.28 \\
         \rowcolor{gray!10} Online-policy & \textbf{43.0} & \textbf{97.58} & \textbf{55.37} \\
    
    \end{tabular}
      \caption{Off- and online learning with VideoScore (w/ TC).}
        \label{tab:policy_learning}
    \end{subfigure}
    \captionsetup{type=table}
    \caption{\textbf{OnlineVPO Analysis Ablation}. We perform ablations on (a) video preference feedback sources. (b) video feedback dimension of online preference learning. (c) preference optimization paradigms. (d) offline- or online-policy exploits the preference feedback.}
    \label{analysis_ablation}
\end{figure*}

\noindent \textbf{Online-policy Superior to Offline-policy.} We argue that the off-policy learning of DPO nature hinders the efficient video preference optimization. To validate this, we utilize the VideoScore to generate preference pairs before training and then align the VDMs with these pairs with standard DPO.  We then compare the performance of this off-policy learning strategy with our OnlineVPO after the same number of optimization steps.  The results, presented in Tab.\ref{analysis_ablation}(d), demonstrate that the online policy approach achieves superior performance than the off-policy counterpart. This improvement arises because the static preference data pairs used in off-policy learning fail to adapt to the evolving distribution of the model's generations during training, leading to undesired feedback. In contrast, OnlineVPO dynamically aligns the preference data with the model's generation capability, ensuring more targeted feedback optimization.

\begin{figure*}[t]
    \centering
    \begin{minipage}[t]{0.45\textwidth}
        \centering
        \setlength{\abovecaptionskip}{0.1cm}
        \includegraphics[width=\linewidth, height=0.6\textwidth]{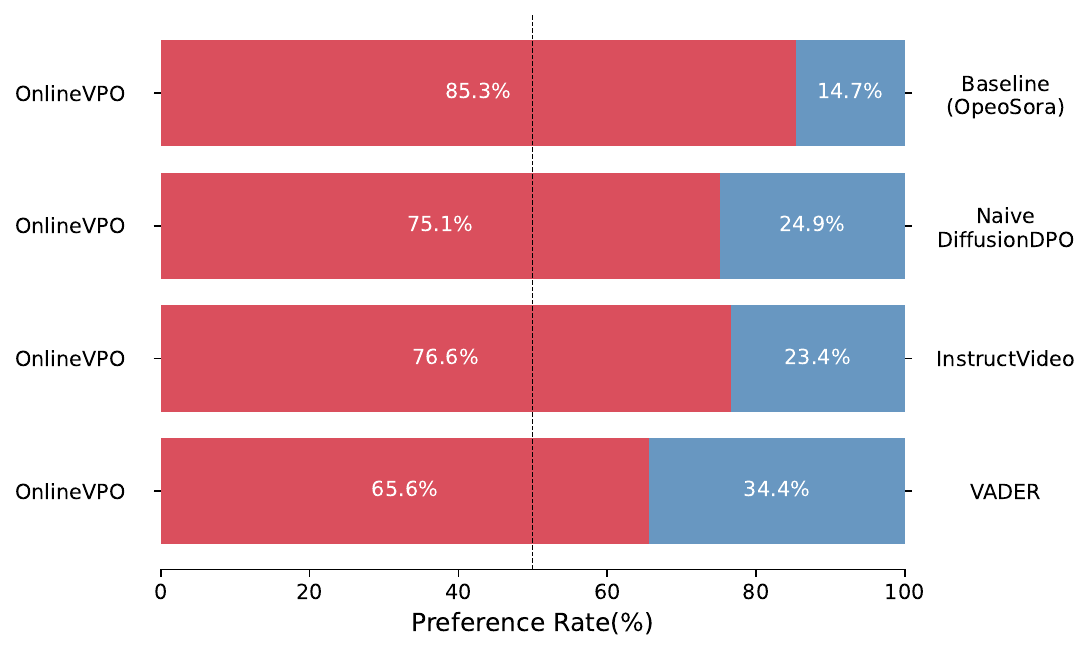}
        \caption{\textbf{User Study} on the performance of the OpenSora optimized by OnlineVPO and other methods.}
        \label{fig:user}
    \end{minipage}
    \hfill 
    \begin{minipage}[t]{0.5\textwidth}
        \centering
        \setlength{\abovecaptionskip}{0.1cm}
        \includegraphics[width=\linewidth]{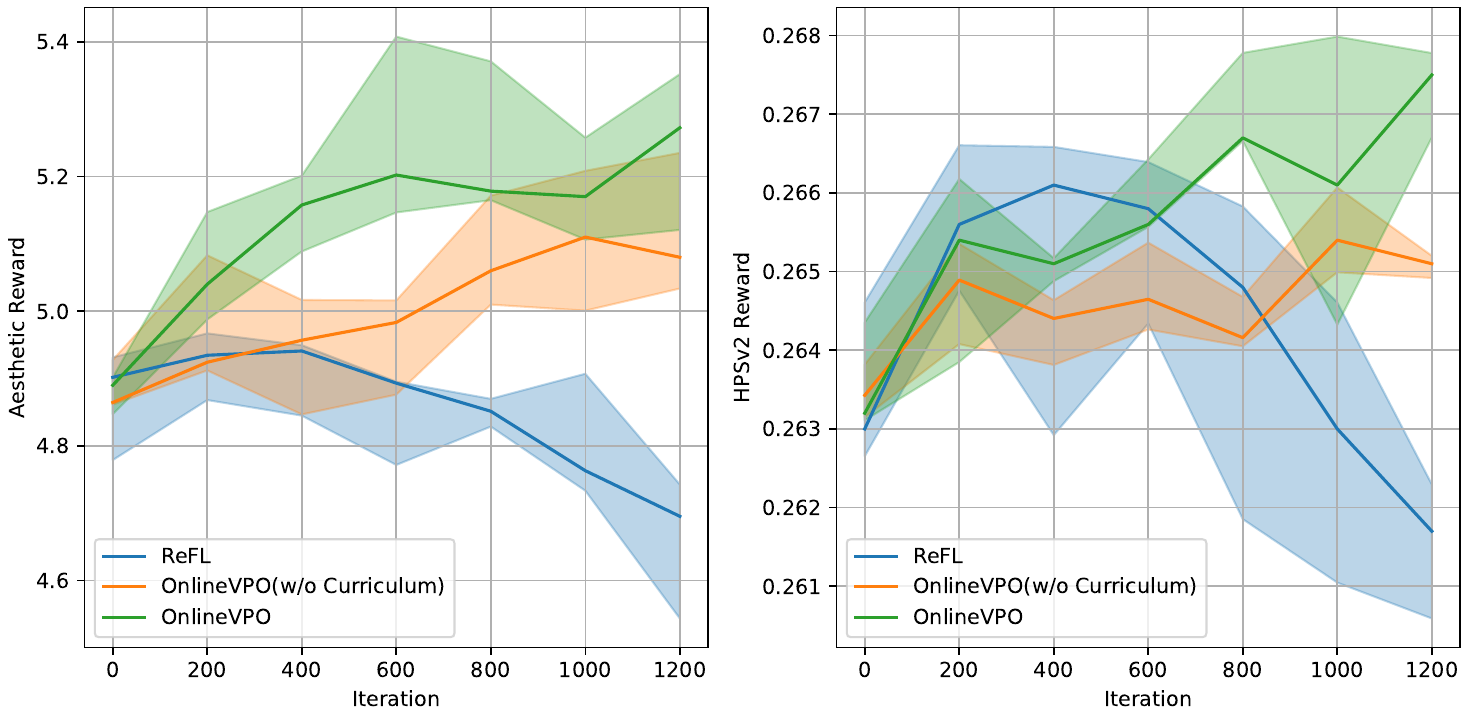}
        \caption{\textbf{Reward Curve} along the training process when adopting the ReFL or OnlineVPO video preference learning paradigm }
        \label{fig:reward_change}
    \end{minipage}
    \vspace{-0.3cm}
\end{figure*}

\noindent \textbf{Reward Learning Efficiency.}
We showcase the reward trends of different preference learning methods during training in Fig.\ref{fig:reward_change}. All methods utilize VideoScore as the feedback source, with HPSv2 and the aesthetic score serving as reward metrics to evaluate the effectiveness of preference optimization. Overall, OnlineVPO demonstrates a stable and consistent increase in rewards throughout training. Additionally, we observe that updating the reference model in a curriculum manner leads to more efficient improvement compared to using a fixed reference model.   In contrast, ReFL initially exhibits a rise in rewards but subsequently experiences a decline, indicating the issue of reward hacking~\cite{rewardhack}—a phenomenon where the reward used for training increases while actual performance deteriorates. We attribute this issue to the unbounded optimization design of ReFL, as shown in Eq.\ref{eq:refl}, which grants the model excessive freedom to explore directions that may bypass the intended optimization objective while still increasing the reward. This problem is likely exacerbated in the video generation domain due to the additional degrees of freedom introduced by the temporal dimension.

\noindent \textbf{Scalability Analysis.}
We further investigate the scalability of different preference optimization methods by comparing the GPU memory requirements of our approach and the ReFL algorithm across various resolutions and frames. The results in Fig.\ref{fig:gpu} demonstrate that our method maintains 25\% of available GPU memory even under demanding settings, such as 720p resolution with 68 frames. In contrast, ReFL is constrained to significantly lower settings, supporting at most 240p resolution with 68 frames or 360p resolution with 17 frames. This stark difference highlights that OnlineVPO is better suited for large-scale video optimization tasks, offering greater scalability and efficiency compared to ReFL.

\section{Conclusion}
We present the first systematic exploration of DPO in the video generation domain and present OnlineVPO, an efficient video preference optimization method. By leveraging video quality assessment models as a reliable alternative to human feedback and the novel online video preference optimization framework, we provide a low-cost, efficient, and scalable solution to enhance video generation performance. Extensive experiments on the open-sourced video generation models demonstrate the superiority of our method.

\paragraph{Acknowledgments} This work is supported by the Hong Kong Research Grants Council - General Research Fund (Grant No.: 17211024).

\clearpage
{
    \small
    \bibliographystyle{ieeenat_fullname}
    \bibliography{main}
}

\end{document}